\documentclass[twocolumn]{article}

\usepackage{amsmath}
\usepackage{amssymb}
\usepackage[ruled]{algorithm2e} 

\usepackage{color}
\usepackage{booktabs} 
\usepackage{algorithmic, multirow, bm}
\usepackage{graphicx}
\usepackage{url}
\usepackage{tikz}
\usepackage{tabularx}
\usepackage[utf8]{inputenc}

\newtheorem{definition}{Definition}
\newtheorem{theorem}{Theorem}
\newtheorem{lemma}{Lemma}
\newtheorem{corollary}{Corollary}
\newenvironment{proof}{ \medskip \noindent{\bf Proof}\ \
}{$\Box$\bigskip}

\newcommand{\prob}{\operatorname{P}}

\newcommand{\pa}{\operatorname{PA}}
\newcommand{\pc}{\operatorname{PC}}

\renewcommand{\emph}{\textit}

\makeatletter
\newcommand*{\indep}{%
  \mathbin{%
    \mathpalette{\@indep}{}%
  }%
}

\newcommand*{\nindep}{%
  \mathbin{
    \mathpalette{\@indep}{/}%
  }%
}

\newcommand*{\@indep}[2]{%
  \sbox0{$#1\perp\m@th$}
  \sbox2{$#1=$}
  \sbox4{$#1\vcenter{}$}
  \rlap{\copy0}
  \dimen@=\dimexpr\ht2-\ht4-.2pt\relax
  \kern\dimen@
  \ifx\\#2\\%
  \else
    \hbox to \wd2{\hss$#1#2\m@th$\hss}%
    \kern-\wd2 %
  \fi
  \kern\dimen@
  \copy0 
}
\makeatother

\begin{document}

\title{A general framework for causal classification\footnote{This is a preprint of an article published in International Journal of Data Science and Analytics. The final authenticated version is available online at: https://doi.org/0.1007/s41060-021-00249-1}}

\author{Jiuyong Li$^1$ \and Weijia Zhang$^1$ \and Lin Liu$^1$ \and Kui Yu$^2$ \and Thuc Duy Le$^1$ \and Jixue Liu$^1$ \\
$^1$ University of South Australia, Australia, Australia \\
$^2$ School of Computer and Information, Hefei University of Science and Technology, China}



\maketitle


\begin{abstract}
In many applications, there is a need to predict the effect of an intervention on different individuals from data. For example, which customers are persuadable by a product promotion? which patients should be treated with a certain type of treatment? These are typical causal questions involving the effect or the \emph{change} in outcomes made by an intervention. The questions cannot be answered with traditional classification methods as they only use associations to predict outcomes. For personalised marketing, these questions are often answered with uplift modelling. The objective of uplift modelling is to estimate causal effect, but its literature does not discuss when the uplift represents causal effect. Causal heterogeneity modelling can solve the problem, but its assumption of unconfoundedness is untestable in data. So practitioners need guidelines in their applications when using the methods.  In this paper, we use causal classification for a set of personalised decision making problems, and differentiate it from classification. We discuss the conditions when causal classification can be resolved by uplift (and causal heterogeneity) modelling methods. We also propose a general framework for causal classification, by using off-the-shelf supervised methods for flexible implementations. Experiments have shown two instantiations of the framework work for causal classification and for uplift (causal heterogeneity) modelling, and are competitive with the other uplift (causal heterogeneity) modelling methods.
\end{abstract}

\textbf{keywords:} Causal effect estimation, Causal heterogeneity, Uplift modelling.



\section{Introduction}

The objective of causal classification is to predict whether a treatment would change an individual's outcome \cite{Fernandez2018_CausalClassification}.
In marketing applications, when the treatment is a promotional advertisement of a product, causal classification is to identify customers likely to purchase the product because of having been shown the advertisement. In medical applications, causal classification is to predict if a treatment would improve a patient's outcome. 

To differentiate causal classification from normal classification, we need to understand the difference between observed and potential outcomes. 
Following the potential outcome model~\cite{Rubin1974_CausalEffect,ImbensRubin2015_Book},  for a treatment $T$, each individual has two potential outcomes, denoted as $Y^1$ and $Y^0$, for the outcomes of the person being treated $T\!=\!1$ and controlled $T\!=\!0$ respectively. At a time point, only one potential outcome can be observed for an individual. For example, if we observe a person buying the product after viewing an advertisement, then $Y\!=\!1\! \mid\! T\!=\!1$ ($Y$ denotes the observed outcome), the potential outcome when $T\!=\!1$ is the same as the observed outcome, i.e. $Y^1 \!=\! 1$, but the other potential outcome, $Y^0$ when $T\!=\!0$, indicating his/her purchase status without viewing the advertisement, is not observed.

Causal classification aims to predict the difference in the potential outcomes $Y^0=0$ and $Y^1=1$, i.e. a change in the outcomes due to the treatment, whereas normal classification predicts whether an individual has the desired (observed) outcome, i.e. $Y^1 =1 \mid T=1$. the observed positive outcome after the treatment regardless of whether this outcome is due to the treatment or not.  

Table~\ref{tab_TypeOfChange} lists the four types of responses to a treatment ($T$ is set to $1$). A positive response means that an individual is positively influenced by the treatment, e.g. a person buys the product as a result of viewing the advertisement. A negative response means that an individual is negatively influenced by the treatment, e.g. a person having planned to buy the product does not buy it since s/he dislikes the advertisement. Nonresponse 0 and nonresponse 1 indicate that the treatment has no impact on an individual, e.g. after having viewed the advertisement, a person having no intention to buy the product still does not buy it (nonresponse 0) and a person having the plan to buy the product buys it (nonresponse 1).

It is difficult to differentiate the four types of responses based on observed outcomes. 
For example, in the observed buying group, we do not know if one  has a positive response or a nonresponse 1 as the observed outcomes are $Y=1$ in both cases, not helping with classifying the responses. 
This reflects the famous quote by John Wanamaker, the pioneer in marketing:  ``Half the money I spend on advertising is wasted; the trouble is I don't know which half." 

Using potential outcomes, causal classification can distinguish the responses 
as indicated in Table~\ref{tab_TypeOfChange}. In the marketing example, a positive response is when a person would not buy the product if s/he did not see the advertisement ($Y^0 = 0$), and s/he has bought the product because of viewing the advertisement ($Y^1 = 1$). Nonresponse 1 is when the person would still buy the product even if s/he did not see the advertisement ($Y^0 = 1$), and s/he has purchased the product by simply using the advertisement as a gateway ($Y^1 = 1$).  
In causal classification, only positive responses are labelled as 1, but in traditional classification, both positive responses and nonresponses 1 are labelled as 1.  
\begin{table}[tb]
	\small
\caption{Types of responses to a treatment. Pos and Neg are abbreviations for Positive and Negative.}
\begin{center}
\begin{tabular}{|c|c|c|c|c|}
\hline
Types of 	& Potential &  Potential & Causal & Normal\\ 
 responses of 	& outcome &  outcome  &  class & class\\
an individual 	&  if $T=0$ 			&  if $T=1$ & label & label\\
\hline
Pos response 	& $Y^0=0$ & $Y^1=1$ & 1 & 1\\
Nonresponse 1		& $Y^0=1$ & $Y^1=1$ & 0 & 1\\
Neg response 	& $Y^0=1$ & $Y^1=0$ & 0* & 0 \\
Nonresponse 0		& $Y^0=0$ & $Y^1=0$ & 0 & 0\\
\hline
\end{tabular}
\end{center}
\label{tab_TypeOfChange}
*{\footnotesize In this table, we only list two classes (1 and 0) for an easy comparison with a typical classification problem. For a multi-class causal classification problem, this should be -1. Please refer to the discussions after Definition~\ref{def-CCProblem}. }
\end{table}

It is challenging to make causal classification based on observational data as it involves counterfactual reasoning. When we observe a purchase by a customer after viewing an advertisement  ($Y^1 = 1$), we need to infer his/her unobserved potential outcome  $Y^0 $, i.e. to answer the counterfactual question: ``Would the customer purchase the product had s/he not viewed the advertisement?'',  to determine whether or not the purchase is a result of viewing the advertisement. 

When data is collected from a randomised experiment, 
an uplift modelling method~\cite{Lo2002_TrueLiftModel,Radcliffe1999_DifferentialResponseAnalysis,Gutierrez2017_CausalUpliftReview,Rzepakowski2012_UpliftMarketing} is used in marketing research to model the causal effect of the treatment as the difference between the probabilities of the observed outcomes in the two groups,  $P(Y \!\mid\! T\!=\!1, \mathbf{X}\!=\!\mathbf{x})\! -\! P(Y \!\mid\! T=\!0, \mathbf{X}\!=\!\mathbf{x})$. The objective of uplift modelling is to estimate causal effect, but assumptions for causal inference have not been discussed in most uplift modelling literature~\cite{DevriendtSurvey2018}. The work in~\cite{Gutierrez2017_CausalUpliftReview} links uplift modelling with causal heterogeneity, but the condition under which uplift is causal effect has not be discussed. An uplift modelling method may not achieve its intended objective when not being used correctly.   


Causal classification can be achieved by causal heterogeneity modelling. Several machine learning methods have been developed recently to discover causal effect heterogeneity~\cite{AtheyImbens2016_PNAS,WagerAthey2018_RF,kunzel2019metalearners}, i.e. to identify the subgroups across which the causal effects of a treatment are different and learn the models for predicting the heterogeneous causal effects across the subgroups. Such a method can be used to predict the causal effect of a treatment on an individual's outcome for causal classification. However, these methods assume that there exists a covariate set satisfying the unconfoundedness assumption~\cite{Rubin1974_CausalEffect,Rosenbaum1983_PropensityScore}. Unfortunately,, the unconfoundedness assumption is untestable in data, and this leaves practitioners wonder how the covariate set should be found.  

This paper makes the following contributions. 
\begin{enumerate}
\item We differentiate causal classification from classification and 
  identify conditions under which the existing uplift modelling methods can be used for causal classification. 
\item We have proposed  an algorithmic framework for causal classification, by linking together normal classification, uplift modelling and graphical causal  modelling. Note that Two Model methods have been used in both uplift and causal heterogeneity modelling literature~\cite{Gutierrez2017_CausalUpliftReview,DevriendtSurvey2018}. Our contribution is to link the methods to specific conditions and a data-driven parent discovery process (covariate selection) to ensure its correctness and efficacy for causal effect estimation.   
\end{enumerate}

\section{Classification, uplift modelling, and causal classification}
We assume a data set with a treatment $T$, an outcome $Y$, a set of all other variables $\mathbf{X}$. For easy notation, we use $y$ for $Y=1$. 
\begin{table}[t]
\caption{Objective functions for classification, uplift modelling and causal classification}
\small
\hspace*{-0.5 cm}
\begin{tabular}{llll}
\hline
\textbf{Normal classification:} maximise likelihood \\
$P(y \mid T=1, \mathbf{X= x})$  \\ 
 \hline
\textbf{Uplift modelling:} maximise difference \\ 
$P(y \mid T=1, \mathbf{X=x}) - P(y \mid T=0, \mathbf{X=x})$  \\ 
 \hline
\textbf{Causal classification:}  estimate conditional causal effect \\
$P(y \mid do(T=1), \mathbf{X=x}) - P(y \mid do(T=0), \mathbf{X=x})$ \\ 
 \hline
\end{tabular}
\label{tab_TypeOfModel}
\end{table}%

In the following, we differentiate the objective functions of classification, uplift modelling, and causal classification, as shown in Table~\ref{tab_TypeOfModel}.

Normal classification is to predict outcome $y$ by maximising the likelihood $P(y \mid T=1, \mathbf{X= x})$. There are many methods to achieve this objective. It makes a probabilistic prediction of an outcome but it does not model $Y$'s change with the change of $T$. 

Uplift modelling aims to maximise the difference in $\prob(y|T=1)$ and $\prob(y|T=0)$ for a given value of $\mathbf{X}$. The difference is modelled explicitly. 
There are some different names in the literature, such as true lift and incremental value~\cite{DevriendtSurvey2018,Gutierrez2017_CausalUpliftReview}, and we use uplift as in the previous surveys. An uplift model is normally built using experimental data to find subgroups which respond differently to a treatment. 
$\mathbf{X}$ is not discussed in the literature and is assumed coming with an experiment. Thinking about an A/B test, customers are randomly selected to be exposed to an advertisement, and this process does not need $\mathbf{X}$ (or only use very few attributes in $\mathbf{X}$). $\mathbf{X}$ is collected separately from an experiment, and contains all attributes related to the individuals as in a normal classification application. However, for the purpose of uplift modelling, there are strong requirements for $\mathbf{X}$. When $\mathbf{X}$ does not satisfy the requirements, the uplift does not indicate the intended causal effect. This is the issue we will address in this paper.     

Causal classification is to estimate the change of $Y$ when an individual takes a treatment $T=1$, and makes use of the conditional causal effect of $T$ on $Y$, i.e. the degree of the change of $Y$ as a result of changing or intervening on $T$ under the condition $\mathbf{X=x}$.  To represent this goal formally, we use Pearl's \emph{do} operator~\cite{Pearl2009_Book}, a notation commonly seen in causal inference literature, to represent an intervention. The \emph{do} operation mimics setting a variable to a certain value (not just passively observing a value) in a real world experiment. The probability given a \emph{do} operation, e.g. $\prob(y \mid do (T = 1))$, indicates the probability of $Y=1$ when $T$ is set to 1, and is different from $\prob(y \mid T = 1)$, the probability of $Y=1$ when observing $T=1$. The objective of causal classification is to estimate conditional causal effect of $T$ on $Y$ given $\mathbf{X=x}$, i.e. $\prob (y\! \mid\! do (T\!=\!1), \mathbf{X\! =x\!})\! -\! \prob (y\! \mid\! do (T\!=\!0), \mathbf{X\!=\!x})$.

Causal heterogeneity modelling aims at estimating conditional causal effects and finding the subgroups in which the causal effects to a treatment deviate from the average causal effect in data~\cite{AtheyImbens2016_PNAS,WagerAthey2018_RF,kunzel2019metalearners}. It is a principled way for causal classification, and some methods are available. However, there methods assume a covariate set satisfying the unconfoundedness assumption~\cite{Rubin1974_CausalEffect}. Unconfoundedness is untestable in data, and this does not help practitioners for using the methods. The conditions identified for uplift modelling in this paper are also applicable to causal heterogeneity modelling. When the conditions are satisfied, we do not distinguish uplift modelling and causal heterogeneity modelling and use uplift (causal heterogeneity) modelling to represent both.    

Based on the objective function of causal classification, we can formally define the causal classification problem in data as follows.
\begin{definition} [Causal classification]
\label{def-CCProblem}
Causal classification is to determine whether a treatment $T$ should be applied (i.e. $do (T= 1)$) to an individual $\mathbf{X}=\mathbf{x}$  by using the test whether conditional causal effect $\prob(y \mid do (T= 1), \mathbf{X=x}) - \prob(y \mid do (T=0), \mathbf{X=x}) > \theta$, where $\theta \ge 0$ is a user specified threshold. 
\end{definition}
The threshold $\theta$ is normally determined based on the application. For example, in personalised advertising, $\theta$ can be determined by the budget of an advertisement campaign and the profit of each successful sale. $\theta$ can be determined by visualisation too. Individuals in a test data set are grouped by deciles of estimated conditional causal effects. The observed differences, $\prob(y \mid T= 1) - \prob(y \mid T=0)$ in the groups are plotted against deciles. It is easy to spot where the observed differences become very small (or negative) and hence $\theta$ is determined in the plot. Alternatively, top $k$ selection can be used instead of setting $\theta$.

Causal classification defined above represents a typical application scenario, such as personalised marketing and medicine where binary decision is required, i.e. a person should be treated or not. However, the theoretical results and methods presented in this paper work for multi-class causal classification problems, including positive response, negative response and non-response (denoted as 1, -1, and 0 in Table~\ref{tab_TypeOfChange}). The multi-class causal classification is easily to be achieved since the output conditional causal effect is continuous and can be split to map multiple classes easily.   

Causal classification is not generally achieved in data since causal effect estimation in data needs strong assumptions~\cite{Rubin1974_CausalEffect,ImbensRubin2015_Book,Pearl2009_Book} which may not be satisfied. Simply speaking, uplift represents the probability difference between treatment (i.e., $T=1$) and control (i.e., $T=0$) in a subgroup of people and can always be observed in data, whereas conditional causal effect indicates the real change of the outcome when the treatment is applied to an individual and may not be estimated in data. A crucial question to answer is when uplifts represent conditional causal effects. 
Our specific objectives in this paper are stated as the following. 

\begin{definition} [Problem statement]
\label{def-ProblemStatement}
Given a data set with a binary treatment variable $T$, a binary or numerical outcome variable $Y$ and a set of other variables $\mathbf{X}$ of any type, this paper aims to (1) identify condition under which causal classification can be achieved by uplift modelling, and (2) develop a framework for causal classification using off-the-shelf classification methods.  
\end{definition}

\section{The conditions for causal classification in data} 
\subsection{Preliminary}
A DAG (directed acyclic graph) $G=(\mathbf{V},\mathbf{E})$ is a directed graph with a set of nodes $\mathbf{V}$ and a set of directed edges $\mathbf{E}$, and no node has a sequence of directed edges pointing back to itself, i.e. there are no loops. If there exists an edge $P\rightarrow Q$ in $G$, $P$ is a parent node of $Q$ and $Q$ is a child node of $P$.  For a node $V\in\mathbf{V}$, we use $\pa(V)$ to denote the set of all its parents. 
A path is a sequence of nodes linked by edges regardless of their directions. A directed path is a path on which all the edges follow the same direction. Node $P$ is an ancestor of node $Q$ if there is a directed path from $P$ to $Q$, and equivalently $Q$ is a descendant of $P$. 
\begin{definition}[Markov condition~\cite{Pearl2009_Book}]
Let $P(\mathbf{V})$ be a probability distribution over the vertices in $\mathbf{V}$ generated by DAG $G=(\mathbf{V}, \mathbf{E})$. $P(\mathbf{V})$ and $G$ satisfy Markov condition if, $\forall V\in\mathbf{V}$, $V$ is conditionally independent of all of its non-descendants given $\pa(V)$.   
\end{definition}

When the Markov condition holds, the joint distribution of $\mathbf{V}$ is factorised as $\prob (\mathbf{V}) = \prod_{V_i\in \textbf{V}} \prob(V_i | \pa (V_i))$. 

\begin{definition}[Faithfulness\cite{Spirtes2000_Book}]
If all the conditional independence relationships in $P(\mathbf{V})$ are entailed by the Markov condition applied to DAG $G=(\mathbf{V},\mathbf{E})$, and vice versa, $P(\mathbf{V})$ and $G$ are faithful to each other.     
\end{definition}

The faithfulness assumption is to ensure that the DAG $G=(\mathbf{V}, \mathbf{E})$ represents all the conditional independence relationships in the joint distribution $P(\mathbf{V})$ and vice versa. 

When we carry out causal inference based on data, the following assumption is essential in addition to the Markov condition and causal faithfulness. 

\begin{definition}[Causal sufficiency~\cite{Spirtes2000_Book}]
For every pair of variables observed in a data set, all their common causes are also observed in the data set.
\end{definition}

A simple understanding of causal sufficiency is that there are no hidden common causes of the variables in a system. 

Given the three assumptions, a DAG learned from data 
is a causal DAG and parents are interpreted as the direct causes of their children.

$d$-Separation as defined below is an important concept to read the conditional independenices/dependencies among nodes from a causal DAG. 

\begin{definition} [$d$-Separation~\cite{Pearl2009_Book}]
A path $p$ in a DAG is $d$-separated by a set of nodes $\mathbf{Z}$ if and only if \\
(1) $\mathbf{Z}$ contains the middle node, $V_k$ of a chain $V_i \to V_k \to V_j$ or $V_i \leftarrow V_k \leftarrow V_j$, or a fork  $V_i \leftarrow V_k \to V_j$ in $p$; and \\
(2) when $p$ contains a collider $V_k$, i.e. $V_i \to V_k  \leftarrow V_j$, none of $V_k$ and its descendants is in $\mathbf{Z}$.
\end{definition}

When nodes $X$ and $Y$ are $d$-separated by $\mathbf{Z}$ in a DAG, we have $(X \indep Y \mid \mathbf{Z})$.

Pearl has invented the $do$-calculus~\cite{Pearl2009_Book} for inferring intervention probabilities using a causal DAG.  

Let $G$ be a causal DAG, $V_1$ and $V_2$ be two variables in $G$. Let $G_{\overline{V_1}}$ represent the  subgraph of $G$ by removing all incoming edges of $V_1$, $G_{\underline{V_2}}$ the subgraph of $G$ by removing all outgoing edges of $V_2$ and $G_{\overline{V_1}, \overline{V_2}}$ the subgraph of $G$ by removing all incoming edges of $V_1$ and $V_2$. $G_{\overline{V_1}\underline{V2}}$ is the subgraph of $G$ by removing all incoming edges of $V_1$ and all outgoing edges of $V_2$. $V_1$ and $V_2$ can be variable sets, the edge removals are then for each variable in the sets. The rules of $do$-calculus are presented as follows.

\begin{theorem} 
	\label {Theorem-Rules}
	[The three rules of $do$-calculus~\cite{Pearl2009_Book}]. $Y$ is the outcome, and $X$, $Z$, $W$ are variables (or variables sets) in DAG $G$. $do(X=x)$ is denoted as $do(x)$ where $x$ is a value of variable $X$.  \\ 
$\textbf{Rule 1}$: Insertion/Deletion of observation: \\$P(y | do (x), z, w) = P(y | do (x), w)$ if $(Y \indep Z \mid X, W)_{G_{\overline{X}}}$; 
\\$\textbf{Rule 2}$: Action/Observation exchange \\ $P(y | do (x), do (z), w) = P(y | do (x), z, w)$ if $(Y \indep Z \mid X, W)_{G_{\overline{X}\underline{Z}}}$; 
\\$\textbf{Rule 3}$: Insertion/Deletion actions \\$P(y | do (x), do (z), w) = P(y | do (x), w)$ if $(Y \indep Z \mid X, W)_{G_{\overline{X}, \overline{Z(W)}}}$ where $Z(W)$ is the set of $Z$ nodes that are not ancestors of any $W$ node in $G_{\overline{X}}$; 
\end{theorem}

Given a causal DAG and an expression of causal effect using $do$ operations. If the probability expression with casual conditioning (i.e. with $do$ operations) is reduced to a standard ($do$-free) probability expression with observed variables by using the above rules one by one, the causal effect is identifiable~\cite{Pearl2009_Book}. 

Conditional causal effect will be used in causal classification. 

\begin{definition}[Conditional causal effect]
Let $T$ be a parent node of $Y$ in a causal DAG, and $\mathbf{X}$ be all other non-descendant variables of $Y$.  The conditional causal effect of $T$ on $Y$ is defined as $(\prob(y \mid do (T= 1), \mathbf{X = x}) - \prob(y \mid do (T=0), \mathbf{X = x}))$.
\end{definition}

 
Conditional causal effect indicates the change of $Y$ resulted from a change of $T$ under condition $ \mathbf{X = x}$. Descendant variables of $Y$ are effect variables of $Y$ and cannot be used as conditions of causal effect on $Y$ since they change with $Y$.
%
%
To estimate conditional causal effects, we firstly need to be sure that it is identifiable, i.e. the probability expression with casual conditioning (i.e. with $do$ operations) is reduced to a standard ($do$-free) probability expression with observed variables. 

\subsection{Conditions for causal classification in data} 

\begin{figure}[!t]
	\centering
	\includegraphics[width=0.25\textwidth]{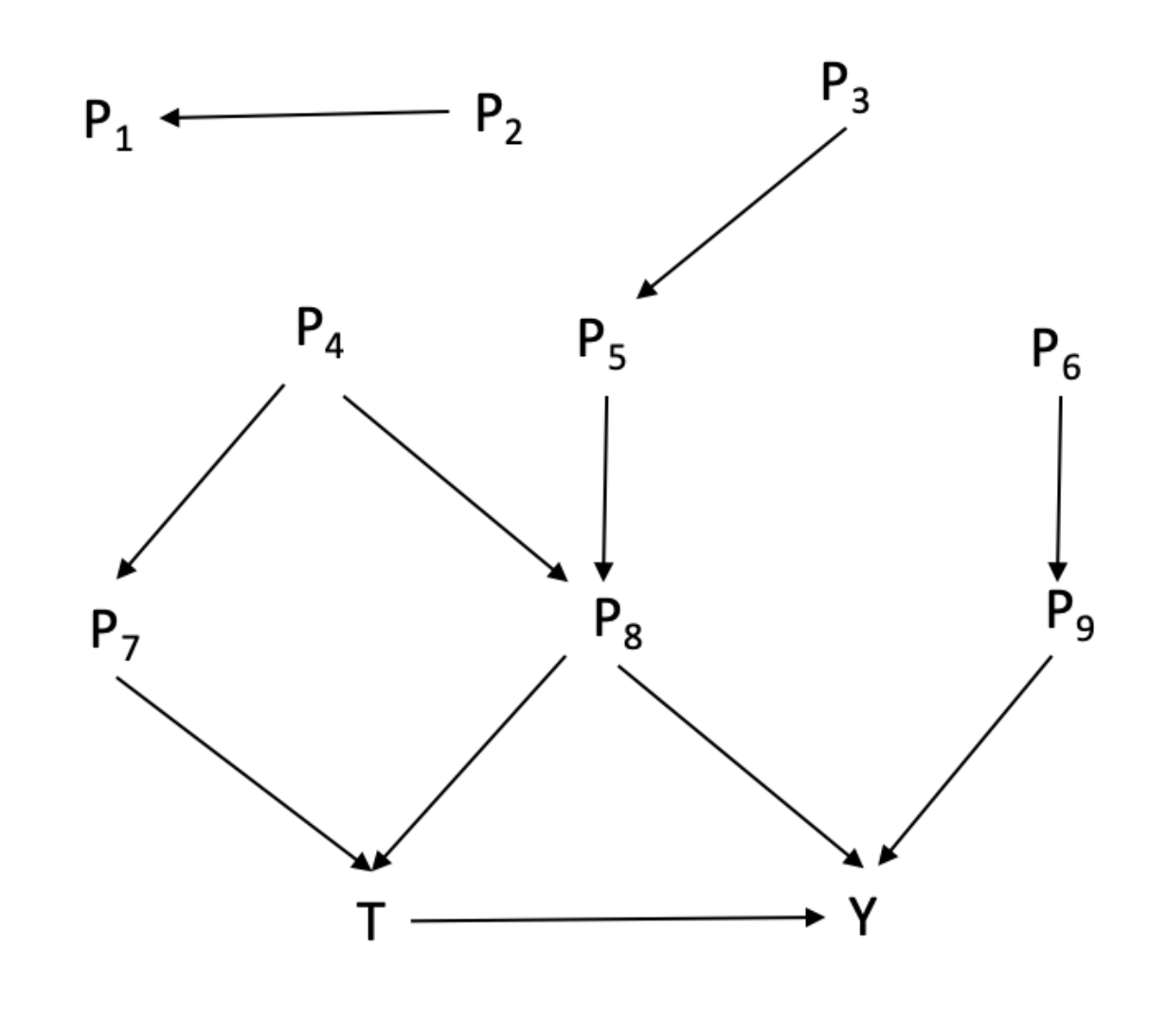} \\
	\caption{A DAG showing a system with pretreatment variables $P_1$ to $P_9$ of $T$ and $Y$. }
	\label{fig-Example}
\end{figure}

We now reduce conditional causal effect to a standard ($do$-free) probability expression with observed variables under a realistic problem setting that 
all variables other than $T$ and $Y$, denoted as $\mathbf{P}$, are pretreatment variables measured before manipulating $T$ and their values are kept unchanged when manipulating $T$. The variables in $\mathbf{P}$ affect the causal effect of $T$ on $Y$ as context since they do not change when the treatment T is manipulated. We also assume that  $Y$ does not have descendants, i.e. the effect variables of $Y$ have not been included in the data set. Note that we do not use $\mathbf{X}$ but $\mathbf{P}$ since we want to indicate the pretreatment condition of our problem setting. An exemplar DAG in the problem setting is shown in Figure~\ref{fig-Example}. There are no descendant nodes of $T$ and/or $Y$ in the DAG. Pretreatment variables can be independent of $T$ and $Y$, such as $P_1$ and $P_2$. All other pretreatment variables in the DAG, $P_3$ to $P_9$, are ancestors of $T$ and/or $Y$. 

This problem setting is realistic as the variables other than $T$ and $Y$ often represent variables describing individuals in a study, e.g. gender, age, and education, which are not affected by $T$ or $Y$. This is often the case in many machine learning problems. 

In order to infer causal effects in data, we should assume that there is no sample selection bias, i.e. all members of the target population have an equal chance to be included in the data set. 

In our problem setting, the conditional causal effect can be reduced to a simple form as follows, given a causal DAG and $Y$'s parents.
\begin{lemma}
	Given a data set containing a set of pretreatment variables $\mathbf{P}$, the outcome $Y$, and the treatment variable $T \in \pa (Y)$, and let $\pa'(Y)=\pa(Y)\backslash\{T\}$, conditional causal effect of $T$ on $Y$ is $(\prob(y \mid do (T=1), \pa'(Y) = \mathbf{p'}) - \prob(y \mid do (T=0), \pa'(Y)=\mathbf{p'}))$.
\label{lemma}
\end{lemma}

\begin{proof}
Let $\mathbf{P} = \pa'(Y) \cup \mathbf{Z} $, and hence $\prob(y \mid do (T=1), \mathbf{P=p}) = \prob(y \mid do (T=1), \pa'(Y) = \mathbf{p'}, \mathbf{Z = z} )$. In DAG $G_{\overline{T}}$ where the incoming edges of node $T$ have been removed, $\mathbf{Z}$ and $Y$ are $d$-separated by $\pa'(Y)$, and hence  $Y \indep \mathbf{Z} \mid \pa'(Y)$. Therefore, $\prob(y \mid do (T=1), \pa'(Y) = \mathbf{p'}, \mathbf{Z = z} ) = \prob(y \mid do (T=1), \pa'(Y) = \mathbf{p'} )$ according to Rule 1 in Theorem~\ref{Theorem-Rules}. 

Similarly, $\prob(y \mid do (T=0), \mathbf{P}) = \prob(y \mid do (T=0), \pa'(Y)=\mathbf{p'})$.

Therefore, the lemma is proved. 
\end{proof}

The above lemma reduces the conditional  set from all pretreatment variables to the parents of $Y$ excluding $T$. The following theorem shows how the conditional causal effect of $T$ on $Y$ is estimated in our problem setting. 

\begin{theorem}
\label{theorem_causalEffect}
Given a data set containing a set of pretreatment variables $\mathbf{P}$, the outcome $Y$, and the treatment variable $T \in \pa (Y)$, and assume that the data set satisfies causal sufficiency. Conditional causal effect of $T$ on $Y$ given $\mathbf{P=p}$, i.e. $\prob(y \mid do (T= 1), \mathbf{P= p}) - \prob(y \mid do (T=0), \mathbf{P = p})$ is equal to $\prob(y \mid T=1, \pa'(Y) = \mathbf{p'})  - \prob(y \mid T=0, \pa'(Y) = \mathbf{p'})$. Hence, causal classification can be resolved by uplift modelling on the projected data set containing $(T, \pa'(Y), Y)$.  
\end{theorem}

\begin{proof}
Firstly, $\prob(y \mid do (T= 1), \mathbf{P= p}) = \prob(y \mid do (T=1), \pa'{(Y)}=\mathbf{p'})$ according to Lemma~\ref{lemma}.

In DAG $G_{\underline{T}}$ where outgoing edges of $T$ have been removed, $\pa'(Y)$ $d$-separate nodes $T$ and $Y$, and hence $Y \indep T \mid \pa'(Y)$. Therefore,  $\prob(y \mid do (T= 1), \pa'(Y) = \mathbf{p'}) = \prob(y \mid T= 1, \pa'(Y) = \mathbf{p'})$ according to Rule 2 in Theorem~\ref{Theorem-Rules}. Therefore, $\prob(y \mid do (T= 1), \mathbf{P= p}) = \prob(y \mid T=1, \pa'(Y) = \mathbf{p'})$.

Similarly, $\prob(y \mid do (T= 0), \mathbf{P= p}) = \prob(y \mid T= 0, \pa'(Y) = \mathbf{p'})$.

Referring to Table~\ref{tab_TypeOfModel}, in this case the objective functions of causal classification and uplift modelling are the same. Therefore, causal classification can be resolved by uplift modelling on the projected data set containing $(T, \pa'(Y), Y)$.  

Therefore, the theorem is proved. 

\end{proof}

The above theorem gives a covariate set for estimating conditional causal effect if we use causal heterogeneity modelling term, and gives a right variable set for uplift modelling. Condition on $\pa'(Y)$, the conditional causal effect can be estimated in data, and uplift estimated in data is the conditional causal effect. In other words,  the uplift is consistent with conditional causal effect in our problem setting. 

In the DAG in Figure~\ref{fig-Example}, for conditional causal effect estimation, only variables $P_8$ and $P_9$ are relevant since they are parent nodes of $Y$ apart from $T$. Note that, this is different from general feature selection since variables $P_3$ to $P_9$ are all correlated to variable $Y$. The correlation between two adjacent nodes may not be larger than that between two non-adjacent nodes. For example, the correlation between $P_9$ and $Y$ may be weaker than the correlation between $P_4$ and $Y$. In general feature selection, $P_4$ is preferred over $P_9$ because of its higher correlation.

Theorem~\ref{theorem_causalEffect}  links causal classification with normal classification since both $\prob(y \mid T=1, \pa'(Y) = \mathbf{p'})$ and   $\prob(y \mid T=0, \pa'(Y) = \mathbf{p'})$ can be estimated by classifiers in data.

In some applications, domain knowledge can be directly used instead of a DAG. Knowing the direct causes of $Y$ is sufficient for causal classification.

\begin{corollary}
\label{corol-directcases}
Let $\{T\} \cup \mathbf{P'}$ be the set of all direct causes of $Y$, causal classification can be achieved by the uplift modelling on data set $(T,  \mathbf{P'}, Y)$.   
\end{corollary}

\begin{proof}
When $\{T\} \cup \mathbf{P'}$ are direct causes of $Y$,  they must be parents of $Y$ in the causal DAG. So, $\pa'(Y) = \mathbf{P'}$. According to Theorem~\ref{theorem_causalEffect}, causal classification can be achieved by uplift modelling on data set $(T,  \mathbf{P'}, Y)$.   
\end{proof}

Theorem~\ref{theorem_causalEffect}  and Corollary~\ref{corol-directcases} establish the conditions under which causal classification can be resolved by uplift modelling. The conditions are also for unbiased estimation of conditional causal effects in data. Under the conditions, causal heterogeneity modelling and uplift modelling are consistent, and they both can be used for causal classification. Hence, in the following, we do not distinguish causal heterogeneity modelling and uplift modelling. 

\section{Framework and algorithm}
In the previous discussions, a DAG is assumed known, However, in a real world application, a DAG is commonly unknown. Therefore, there are two key components in our causal classification framework, finding $\pa(Y)$ and building classification models on the projected data set.
We can obtain $\pa(Y)$ by using domain knowledge (i.e. direct causes of $Y$), or by learning from data. 

In this paper, we present a framework where users can assemble their own causal classification system using  off-the-shelf machine learning methods. Note that the two model approach described below is not new and has been used in uplift modelling and causal heterogeneity modelling~\cite{Gutierrez2017_CausalUpliftReview}. Here, we put it in a framework with a data driven covariate selection process (finding parents) to ensure the soundness of uplifting modelling and hence the framework is new.  


\subsection{Finding parents of $Y$ in data}
\label{sec-PC}

When we do not know the causes of $Y$, finding $\pa(Y)$ from data is the first step for causal classification. One straightforward way is to learn an entire causal DAG from data and then to read $\pa(Y)$ from the DAG, However, learning an entire DAG is computationally expensive or intractable with high dimensional data. Furthermore, it is often unnecessary and wasteful to find the entire DAG when we are only interested in the local structure around $Y$. 

Local structure discovery~\cite{Aliferis2010a,CausalSurvey2020} fits our purpose better. Currently there are mainly two types of local structure discovery methods, one for identifying $\pc(Y)$, the set of Parents (direct causes) and Children (direct effects) of the target $Y$; and one for discovering $MB(Y)$, the Markov Blanket of $Y$, i.e. the parents, children and spouses (the parents of the children) of $Y$. Discovering $\pc(Y)$ is sufficient in our problem setting, $Y$ does not have descendants, i.e. $\pc(Y)=\pa(Y)$. 
Several algorithms have been developed for discovering $\pc(Y)$, such as MMPC (Max-Min Parents and Children)~\cite{Tsamardinos2006_MMPC} and  HITION-PC~\cite{Aliferis2003_Hiton}. 
These algorithms use the framework of constraint-based Bayesian network learning and employ conditional independence tests for discovering $\pc(Y)$. 

\begin{algorithm}

	\caption{Causal Classification by the Two Model approach (CCTM)}
	\label{alg-CCTM}
	
	\hspace{0.5cm}/*---Training---*/
	
	{\bf{Input}}: Data set $D$ containing treatment variable $T$, pretreatment variables $\mathbf{P}$ and outcome variable $Y$.  \\
	{\bf{Output}}: Two models $(M_{T=1}, M_{T=0})$ .
	
	\begin{algorithmic}[1]
		\STATE call a local PC algorithm to find $\pa(Y)$ 
		\STATE let $\pa'(Y) = \pa(Y) \backslash T$ 
		\STATE project data set $D$:$(T, \mathbf{P}, Y)$ to $D'$:$(T, \pa'(Y), Y)$
		\STATE split data set $D'$ to $D_1 \mid T=1$ and $D_0 \mid T=0$
		\STATE call a classification method to build a classifier $M_{T=1}$ on $D_1$
		\STATE call a classification method to build a classifier $M_{T=0}$ on $D_0$
		\STATE output $(M_{T=1}, M_{T=0})$
	\end{algorithmic}
	
	\vspace{0.2cm}
	\hspace{0.5cm}/*---Prediction---*/
	
	{\bf{Input}}: Model pair $(M_{T=1}, M_{T=0})$, $\pa'(Y)$,  test data set $D_{T}$ without treatment assignment and outcome, and a user specified threshold $\theta$ \\
	{\bf{Output}}:  $D_T$:$(\hat{T}, \mathbf{P}, CE)$ where $\hat{T}$ contains treatment assignment and $CE$ contains estimated conditional causal effects.
	
	\begin{algorithmic}[1]
		\STATE project data set $D_T$:$(\mathbf{P})$ to $D_T'$:$(\pa'(Y))$ 
		\FOR {each $r \in D_T'$}
		\STATE let $\prob(y \mid T=1, r) = M_{T=1}(r)$
		\STATE let $\prob(y \mid T=0, r) = M_{T=0}(r)$
		\STATE {\textbf{\textit{if}} $\delta = \prob(y \mid T=1, r) - \prob(y \mid T=0, r)) > \theta$  \textbf{\textit{then}} let $t=1$} 
		\STATE {\textbf{\textit{else}} let $t=0$} 
		\STATE add record $(\hat{T}=t, r, CE = \delta)$
		\ENDFOR
		\STATE output $D_T$:$(\hat{T}, \mathbf{P}, CE)$
	\end{algorithmic}
\end{algorithm}

\subsection{Two Model approach}

Our framework builds a causal classification model and conducts classification using the following Two Model approach. 

\begin{definition}[Two Model approach]
\label{def-two-model}
Given a data set $D$ and assume causal sufficiency and $T \in \pa (Y)$. Let $M_{T=1}$ and $M_{T=0}$ be two classifiers built with $D_{\Pi (\pa'(Y)) \wedge (T=1)}$ and $D_{\Pi (\pa' (Y)) \wedge (T=0)}$ respectively, i.e.  the sub-data sets projected from $D$ to $\pa'(Y)$ for $T=1$ and $T=0$ respectively. 
The test for causal classification
in Definition \ref{def-CCProblem} can be achieved by $\prob(y \mid M_{T=1} (\mathbf{P'=p'})) - \prob(y \mid M_{T=0}(\mathbf{P'=p'})) > \theta$, where $\mathbf{P'} = \pa'(Y)$.  
\end{definition}

Based on the proposed framework, we present the Causal Classification by the Two Model approach (CCTM) in Algorithm~\ref{alg-CCTM}. The training phase of CCTM is to build two classifiers using variables in $\pa'(Y)$ in the two sub data sets containing $T=1$ and $T=0$ respectively. Any classification method, such as decision tree or SVM can be plugged in to build the classifiers. In the prediction phase, the trained classifier pairs $M_{T=1}$ and $M_{T=0}$ predict whether a treatment will lead to a positive response (effect) or not. Line 1 of the prediction phase projects the test data set to contain the same variables in $\pa'(Y)$ in order to use the two classifiers to estimate $\prob(y \mid T=1, \mathbf{P'= p'})$ and $\prob(y \mid T=0, \mathbf{P'= p'})$ respectively for an individual. If the difference in the probabilities (estimated conditional causal effect) is larger than $\theta$, the individual is predicted to have a positive response and should be treated. Otherwise, the treatment should not be applied to the individual.%

\section{Experiments}

This section serves as a demonstration that the proposed framework works with off-the-shelf methods. In Section~\ref{sec_parentDisocvery}, we show that the parents of $Y$ can be discovered in data when the conditions for Theorem~\ref{theorem_causalEffect} are satisfied, and the discovery accuracy and time efficiency are satisfactory. In Section~\ref{sec_causalEffectestimation}, we use synthetic data sets to show the efficacy of two instantiations of the framework. To demonstrate the usefulness of Theorem~\ref{theorem_causalEffect}, for each of the methods (the two CCTM instantiations, other uplift  and causal heterogeneity modelling methods), we compare its performance when using parent variables with its performance when using all variables. The results show that using parents is useful for all methods. In Section~\ref{sec_Exp-Realworld}, we show that two instantiations of the framework work in real world data sets, in comparison with some existing methods, and demonstrate that the two CCMT instantiations perform competitively with other methods.      

\subsection{Parent discovery in data}
\label{sec_parentDisocvery}
In this section, we show how to use local structure learning algorithms MMPC and HITON-PC~\cite{Aliferis2010a} to achieve the first step in the framework: to find the parents of the outcome variable. We also demonstrate their performance for parent discovery. Their implementations are from the Causal Explorer package~\cite{Statnikov2010_CausalExplorer}, and $G^2$ test (with significance level 0.01) is used for conditional independence test. For the conditional independence tests, the maximum size of a conditioning variable set is 3 for both algorithms. The experiments are done on a PC with Intel(R)  i5-8400 and 16GB memory. 

We will use data sets which have known parents for evaluation, i.e. we use the known parents as the ground truth to evaluation the results of parent discovery. Four benchmark Bayesian networks (BNs), CHILD~\cite{Cowell2006_bnCHILD}, ALARM~\cite{Beinlich1989_bnALARM}, PIGS~\cite{Jensen1997_thesis_bnPIGS}, and GENE~\cite{Spellman1998_bnGENE} (\url{www.bnlearn.com/bnrepository}) containing 20, 37, 441 and 801 variables respectively, are used to generate the evaluation data sets. For each BN, we generate data sets  with 500, 1000, and 5000 samples respectively. For each sample size, we generate a group of 10 data sets, so in total 120 data sets are generated for the four BNs. We make use of nodes having no descendants as the treatment or outcome variable, to be consistent with our problem setting, i.e. $T$ and $Y$ have no descendants and all other variables are pretreatment. 

The quality of parent discovery is measured by the average precision, recall, and F1 score of the discovered parents against the known parents in each data set. The average precision, recall, and F1 score are reported in Table~\ref{tb9-11}. In most cases, the algorithms produce accurate results. For the data sets with 5000 samples, both MMPC and HITON-PC achieve perfect results with 100\% precision and recall. This shows that if the data set is large, the parent discovery can be accurate. 

\begin{figure}[t]
	\centering
	\begin{tabular}{ll}
	\includegraphics[width=0.22\textwidth]{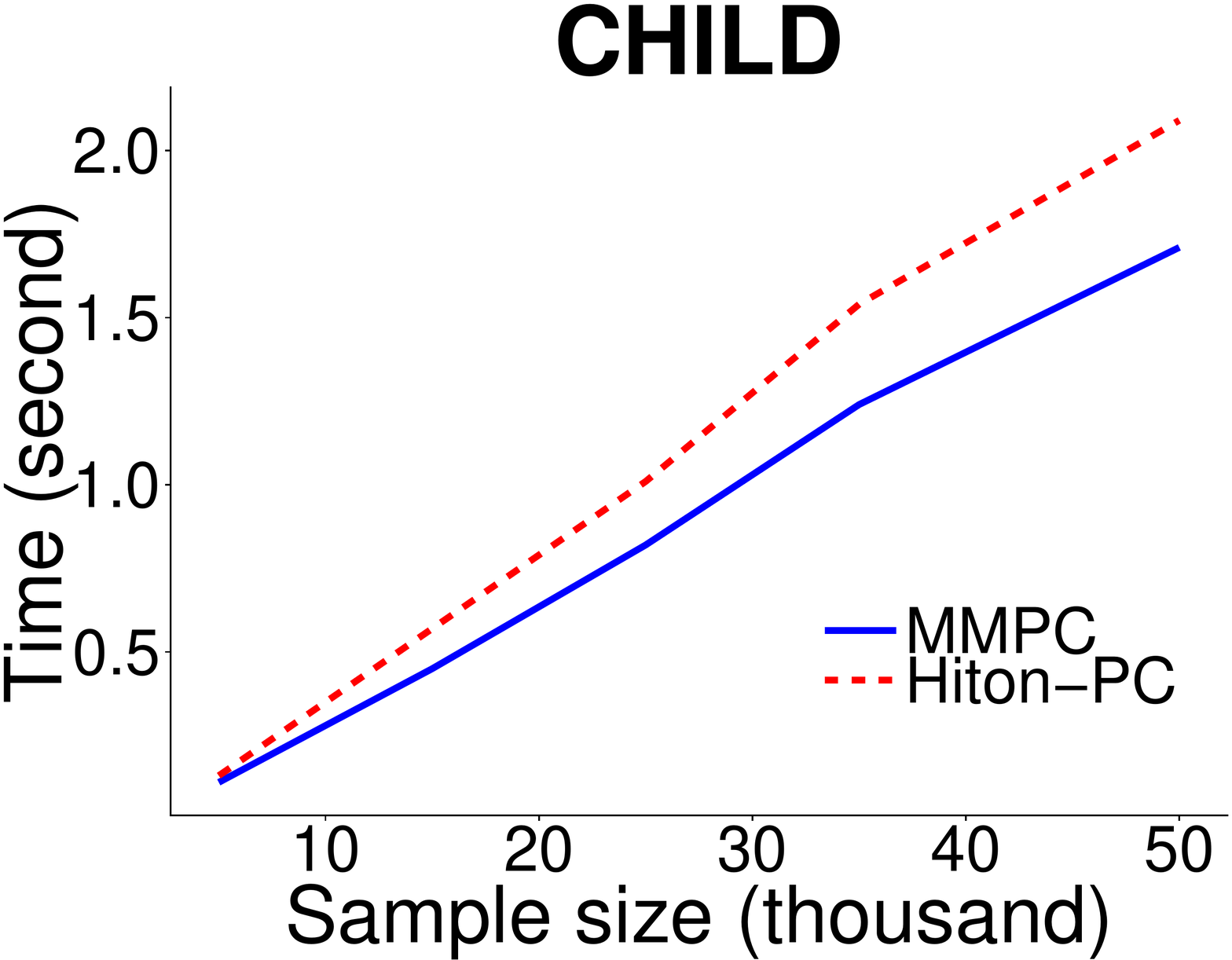} &
	\includegraphics[width=0.22\textwidth]{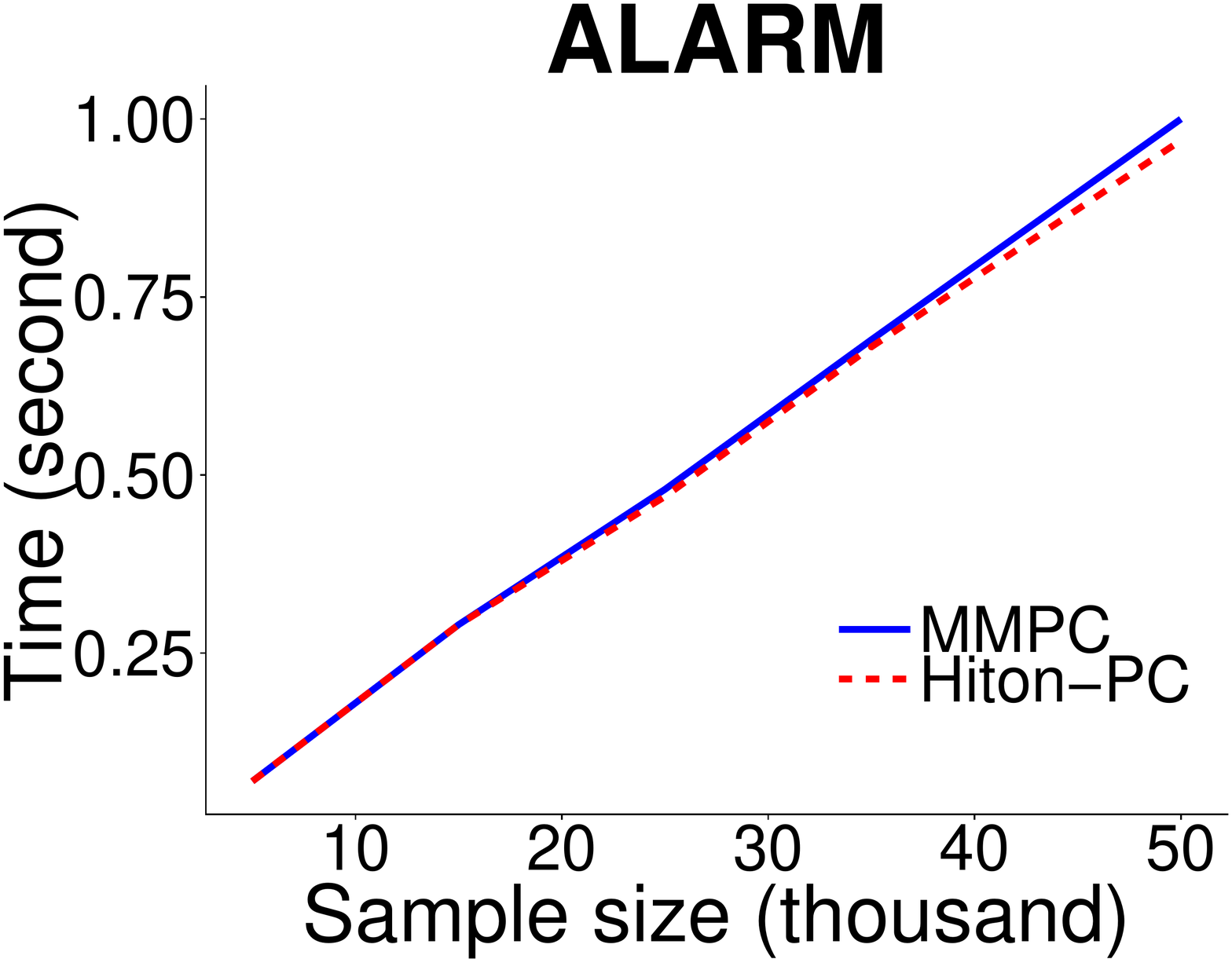}\\
	\includegraphics[width=0.22\textwidth]{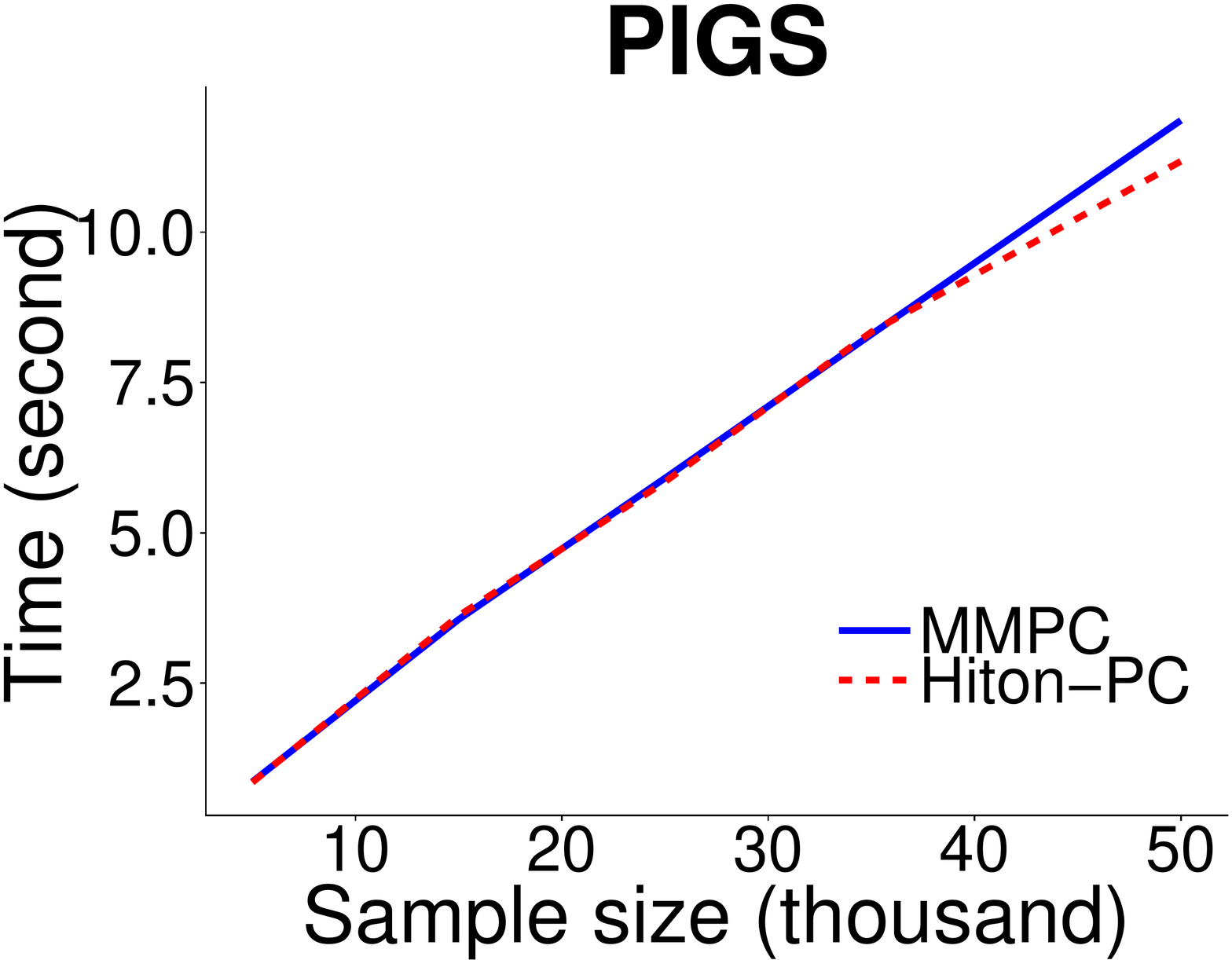} &
	\includegraphics[width=0.22\textwidth]{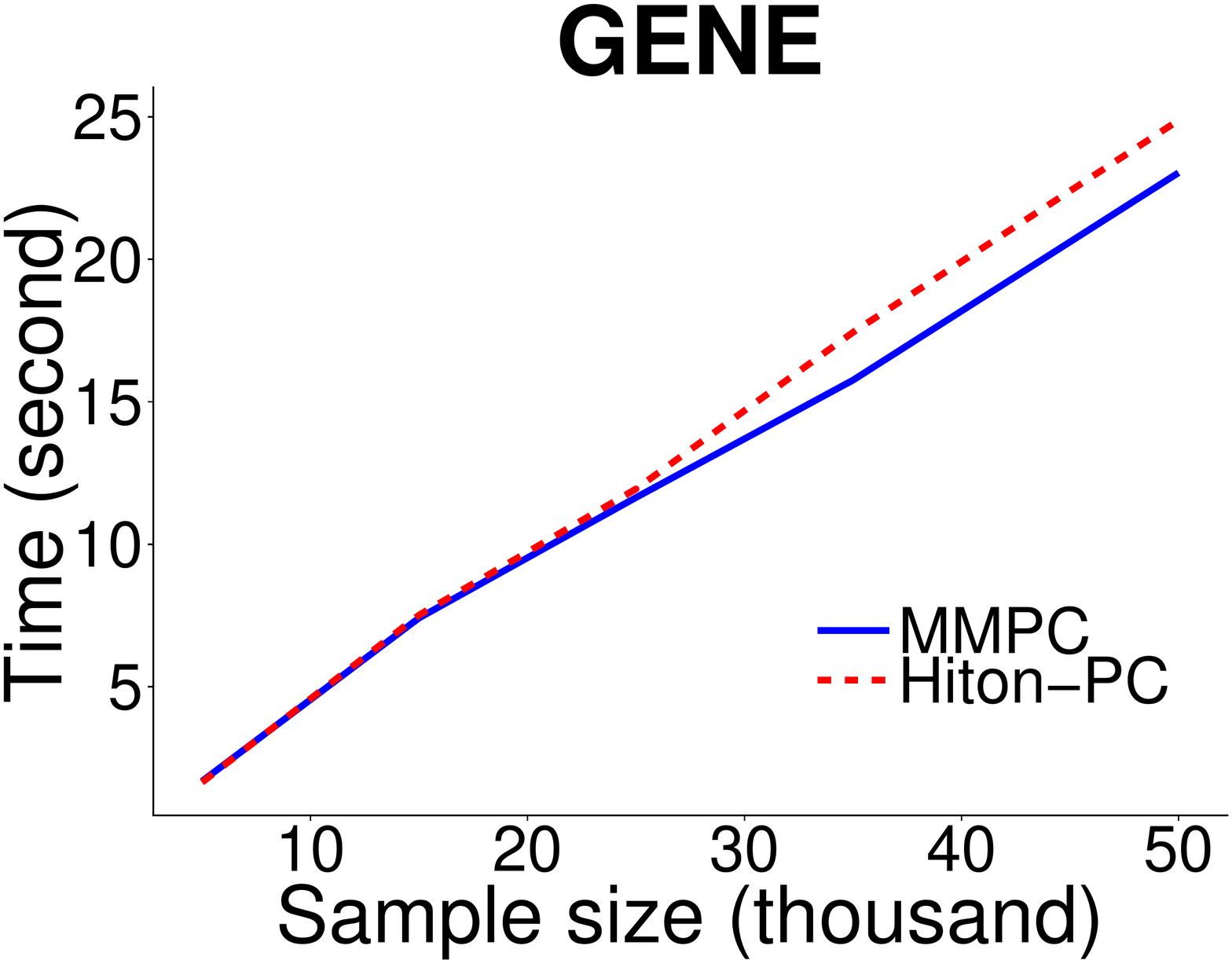}
	\end{tabular}
	\caption{Scalability of MMPC and HITON-PC}
	\label{scalability}
\end{figure}

To show the time efficiency and scalability of the local structure learning algorithms MMPC and HITON-PC, we generate data sets with 5 K, 15 K, 25 K, 35 K and 50 K samples respectively. As shown in Figure~\ref{scalability}, both MMPC and HITON-PC are fast and scalable to the size of data sets.  

In the proposed CCTM algorithm, MMPC is used because it is slightly faster than HITON-PC.

\begin{table}
\centering
\footnotesize
\caption{Quality of parent discovery}
\begin{tabular}{|l|c|l|l|l|l|}
\hline
\multicolumn{1}{|c|}{BN}                     & Size               & \multicolumn{1}{c|}{Alg} & \multicolumn{1}{c|}{Precision} & \multicolumn{1}{c|}{Recall} & \multicolumn{1}{c|}{F1 score} \\ \hline
            \multirow{6}{*}{\rotatebox{270}{CHILD}}                      &        \multirow{2}{*}{500}               & MMPC                           & 97.0$\pm$0.2                     & 90.0$\pm$0.1                   & 92.0$\pm$0.1              \\ \cline{3-6} 
                                             &                       & HITON                       & 97.0$\pm$0.2                      & 90.0$\pm$0.1                   & 92.0$\pm$0.2              \\ \cline{2-6} 
                                             &    \multirow{2}{*}{1000}                   & MMPC                           & 100$\pm$0.0                      & 100$\pm$0.0                   & 100$\pm$0.0             \\ \cline{3-6} 
                                             &                       & HITON                       & 100$\pm$0.0                      & 100$\pm$0.0                   & 100$\pm$0.0             \\ \cline{2-6} 
                                            &            \multirow{2}{*}{5000}            & MMPC                           & \textbf{100$\pm$0.0}             & \textbf{100$\pm$0.0}          & \textbf{100$\pm$0.0}      \\ \cline{3-6} 
                                             &                       & HITON                       & \textbf{100$\pm$0.0}             & \textbf{100$\pm$0.0}          & \textbf{100$\pm$0.0}      \\ \hline
{\multirow{6}{*}{\rotatebox{270}{ALARM}}}                       &           \multirow{2}{*}{500}             & MMPC                           & 60.0$\pm$0.3                     & 90.0$\pm$0.3                   & 71.0$\pm$0.2               \\ \cline{3-6} 
\multicolumn{1}{|c|}{}                       &                       & HITON                       & 60.0$\pm$0.3                      & 90.0$\pm$0.2                   & 71.0$\pm$0.2               \\ \cline{2-6} 
\multicolumn{1}{|c|}{}                       &    \multirow{2}{*}{1000}                   & MMPC                           & 92.0$\pm$0.0                     & 100$\pm$0.1                   & 95.0$\pm$0.1              \\ \cline{3-6} 
\multicolumn{1}{|c|}{}                       &                       & HITON                       & 92.0$\pm$0.0                      & 100$\pm$0.1                   & 95.0$\pm$0.0               \\ \cline{2-6} 
\multicolumn{1}{|c|}{}                       &    \multirow{2}{*}{5000}                   & MMPC                           & \textbf{100$\pm$0.0}             & \textbf{100$\pm$0.0}          & \textbf{100$\pm$0.0}      \\ \cline{3-6} 
\multicolumn{1}{|c|}{}                       &                       & HITON                       & \textbf{100$\pm$0.0}             & \textbf{100$\pm$0.0}          & \textbf{100$\pm$0.0}      \\ \hline
            \multirow{6}{*}{\rotatebox{270}{PIGS}}                 &        \multirow{2}{*}{500}               & MMPC                           & 91.0$\pm$0.0            & 100$\pm$0.1                   & 95.0$\pm$0.0               \\ \cline{3-6} 
                                             &                       & HITON                       & 92.0$\pm$0.0                      & 100$\pm$0.1                   & 95.0$\pm$0.0               \\ \cline{2-6} 
                                             &           \multirow{2}{*}{1000}            & MMPC                           & 100$\pm$0.0                      & 100$\pm$0.0                   & 100$\pm$0.0               \\ \cline{3-6} 
                                             &                       & HITON                       & 100$\pm$0.0                      & 100$\pm$0.0                   & 100$\pm$0.0               \\ \cline{2-6} 
                                             &         \multirow{2}{*}{5000}              & MMPC                           & \textbf{100$\pm$0.0}             & \textbf{100$\pm$0.0}          & \textbf{100$\pm$0.0}      \\ \cline{3-6} 
                                             &                       & HITON                       & \textbf{100$\pm$0.0}             & \textbf{100$\pm$0.0}          & \textbf{100$\pm$0.0}      \\ \hline
           \multirow{6}{*}{\rotatebox{270}{GENE}}       &         \multirow{2}{*}{500}              & MMPC                           & 76.0$\pm$0.1                      & 95.0+0.2                   & 82.0+0.1               \\ \cline{3-6} 
                                             &                       & HITON                       & 76.0$\pm$0.1                      & 95.0$\pm$0.2                   & 83.0$\pm$0.1               \\ \cline{2-6} 
                                             &        \multirow{2}{*}{1000}                & MMPC                           & 72.0$\pm$0.0                     & 100$\pm$0.2                   & 82.0$\pm$0.1              \\ \cline{3-6} 
                                             &                       & HITON                       & 83.0$\pm$0.0                      & 100$\pm$0.2                   & 89.0$\pm$0.1               \\ \cline{2-6} 
                                             &           \multirow{2}{*}{5000}            & MMPC                           & \textbf{100$\pm$0.0}             & \textbf{100$\pm$0.0}          & \textbf{100$\pm$0.0}      \\ \cline{3-6} 
                                             &                       & HITON                       & \textbf{100$\pm$0.0}             & \textbf{100$\pm$0.0}          & \textbf{100$\pm$0.0}      \\ \hline
\end{tabular}
\label{tb9-11}
\end{table}

\subsection{Evaluation on synthetic data sets}
\label{sec_causalEffectestimation}

We demonstrate the efficacy of the CCTM framework by using two instantiations of it. We use two popular classifiers, SVM and Random Forest (RF) to instantiate the proposed causal classification framework to two algorithms, denoted as CCTM-SVM and CCTM-RF respectively. The implementations of RF and SVM are from \\  \url{https://cran.r-project.org/web/packages/randomForest/index.html} and \url{https://www.csie.ntu.edu.tw/~cjlin/libsvm/ respectively}. Default parameters are used. 

We will demonstrate that the instantiated algorithms CCTM-SVM and CCTM-RF perform well in causal classification, and that Theorem~\ref{theorem_causalEffect} is generally applicable to uplift (and causal heterogeneity) modelling methods. In this evaluation, for each method, we compare its performance when using parent variables with its performance when using all variables.   

To benchmark the instantiated algorithms, we also run some well known uplift (and causal heterogeneity) modelling methods, including Uplift Random Forests (Uplift RF)~\cite{Guelman2015_UpliftRF}, Uplift Causal Conditional Inference Forests (Uplift CCIF)~\cite{Guelman2014_CCIF}, t-Statistics Tree~\cite{Su2009_RecursivePartitioning}, CausalTree~\cite{AtheyImbens2016_PNAS},  and the X-Learner~\cite{kunzel2019metalearners}. We also compare the instantiations with two treatment responder classification methods under causal effect monotonicity~\cite{kallusCausalclassifiction19}: RespSVM-Linear and RespLR-Gen. In our experiments, we use the implementations of the methods from authors' or commonly used packages: Uplift RF and Uplift CCIF from \url{https://cran.r-project.org/web/packages/uplift/index.html}, t-Stats Tree and Causal Tree from \url{https://github.com/susanathey/causalTree}, X-Learner from \url{https://github.com/soerenkuenzel/causalToolbox}, and RespSVM-Linear and RespLR-Gen from \url{https://github.com/CausalML/classifying-responders}. Default parameters are used for all methods except RespSVM-Linear and RespLR-Gen. The parameter on the regularization term for RespSVM-Linear is selected by 5-fold cross validation on each training data set from the set (0.001, 0.01, 0.1, 1, 10, 100, 1000). RespLR-Gen is a neural network with no hidden layer learning by the Adam optimiser using 200 epochs as in~\cite{kallusCausalclassifiction19}.

\begin{figure}[t!]
	\centering
	\begin{tabular}{ll}
    \includegraphics[width=0.22\textwidth]{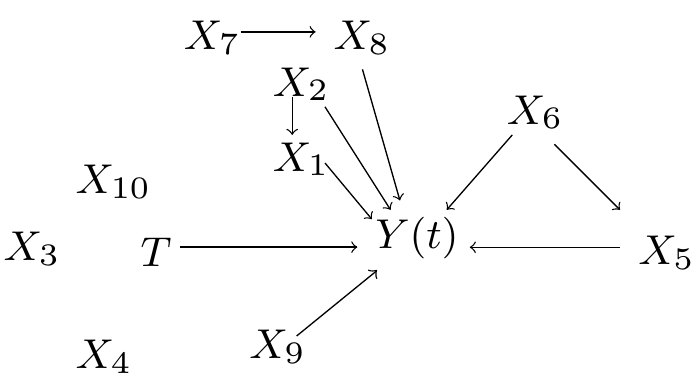}&
    \includegraphics[width=0.22\textwidth]{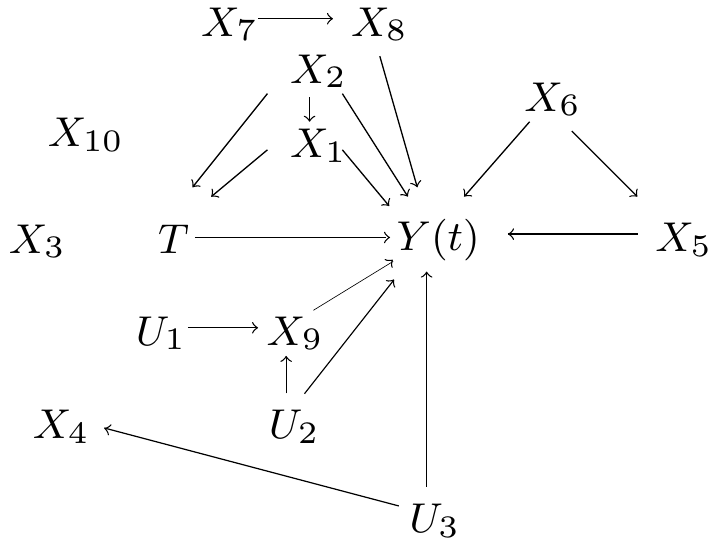}
	\end{tabular}
	\caption{Causal DAGs for synthetic data generation. Left: for Group 1 data sets; Right: for Group 2 data sets.}
	\label{simulation}
\end{figure}

We will need data sets with known ground truth (true uplifts or conditional causal effects) for evaluation. Two groups of simulation data sets (Group 1 and Group 2) are generated following the work in~\cite{ConfounderSelection2018}. The generation program is at \url{https://cran.r-project.org/web/packages/CovSelHigh/index.html}. The causal DAGs used for generating the data sets are shown in Figure~\ref{simulation}. Each group contains 10 data sets, and each data set has 10,000 samples and 102 variables. $T$ and $Y$ are binary. Apart from $X_1$ to $X_{10}$ in the DAGs, other 90 variables which are irrelevant to $T$ and $Y$ are included to simulate real world situations. 100 variables are drawn from a mixture of continuous and binary distributions.  
where the left graph (for generating Data set 1) is to simulate the situation when causal sufficiency is satisfied, and the right graph (for generating Data set 2) is to simulate the situation when hidden variables exist. When generating a data set in Group 2, after obtaining the data set based on the right structure in Figure \ref{simulation}, we remove the columns for variables $U_1$,$U_2$ and $U_3$ from the data set to simulate latent variables. The DAG underlying Group 2 data sets does not cause a major problem for the CCTM methods although the casual sufficiency assumption is not satisfied. In the DAG, the two paths with hidden variables do not link to $T$ and hence hidden variables do not cause a bias in causal effect estimation. $X_4$ is not a parent of $Y$ but is a good proxy of $U_3$ (which is the parent of $Y$). Some effect of $U_2$ on $Y$ (not via $X_9$) is missed, and this results in an uncertainty for all methods.

For evaluation of causal classification, a half of data set is used for training models and another half is used to test the accuracy. Threshold $\theta$ is set to 0 to separate positive causal effect from zero or negative causal effect. A prediction is correct if the treatment assignment is the same as the assignment based on the ground truth causal effect for data generation.

The accuracies of all methods are presented in Table~\ref{simulation_table}. Each method is run by using all variables and $\pa'(Y)$ respectively and the two accuracies are compared. T-test is conducted with the null hypothesis that there is no difference between two accuracies achieved using all variables and $\pa'(Y)$. The confidence level of the t-test is set at 95\%, and significant results are marked by *.  Using parent variables consistently achieves higher accuracy than using all the variables for all methods. Most differences are statistically significant. This means that our theoretical result, i.e. Theorem~\ref{theorem_causalEffect}, improves all uplift (causal heterogeneity) modelling methods. 

Table~\ref{simulation_comparison} presents the results of t-tests of accuracies between a pair of a CCTM method and a comparison method. Using off-the-shelf package without parameter tuning, the two instantiations achieve mostly better performance in comparison with other methods (17/28). They perform worse than other in three cases. Considering that the comparison methods are tailor designed (some are asymptotic) solutions, the performance CCTM methods is very good.   


\begin{table}
\footnotesize
\centering
	\caption{Accuracies of causal classification. All methods are run with two versions, using all variables (original) and $\pa'(Y)$ respectively, and the higher accuracy between the two is marked in bold. * indicates that the accuracy difference between the two versions of a method is significant at 95\% confidence level. Parent nodes improve the accuracies generally. }
	\begin{tabular}[width  = \textwidth]{ | l| c| c | c | c | c| }
		\hline 
Method  & Strategy & Group 1 & Group 2 \\

\hline
		Two Model RF & All & 71.2$\pm$1.2  & 78.1$\pm$1.3 \\

CCTM RF & $\pa'(Y)$ & \textbf{84.1$\pm$0.8}*	& \textbf{88.3$\pm$0.6}* \\

\hline

Two Model SVM & All & 84.4$\pm$0.9  & 87.9$\pm$1.2  \\

CCTM SVM & $\pa'(Y)$ &  \textbf{84.8$\pm$1.5}*		& \textbf{88.6$\pm$1.1}* \\
\hline
		
		Causal Tree &  All & 80.0$\pm$0.5 & 73.4$\pm$1.3   \\
		Using $\pa'(Y)$ & $\pa'(Y)$   &\textbf{81.2$\pm$0.5}* 		 & \textbf{79.3$\pm$0.9}* \\
		\hline
		
		RespLR-Gen &  All & 80.2$\pm$2.3 & 88.1$\pm$1.3   \\
  Using $\pa'(Y)$ & $\pa'(Y)$   &\textbf{83.2$\pm$1.1}* 		 & \textbf{88.9$\pm$0.9}* \\
\hline

		RespSVM-Linear &  All & 83.0$\pm$1.9 & 89.3$\pm$1.1   \\

 Using $\pa'(Y)$ & $\pa'(Y)$   &\textbf{83.2$\pm$0.8} 		 & \textbf{89.5$\pm$1.0} \\
\hline

		t-Stats Tree & All & 34.5$\pm$2.1& 11.6$\pm$1.3  \\	
		Using $\pa'(Y)$ & $\pa'(Y)$ &  \textbf{74.6$\pm$0.8}*  & \textbf{81.7$\pm$5.1}*  \\
		\hline
		
		Uplift CCIF & All & 77.4$\pm$1.0  & 89.0$\pm$1.3  \\		
		Using $\pa'(Y)$ & $\pa'(Y)$ & \textbf{78.8$\pm$1.3}*		 & \textbf{89.2$\pm$0.6} \\
		\hline
		
		Uplift RF & All & 77.3$\pm$1.4 & 89.2$\pm$1.0 \\
		Using $\pa'(Y)$ & $\pa'(Y)$ & \textbf{78.9$\pm$1.6}*		 & \textbf{89.3$\pm$1.0} \\
		\hline
		
			X-Learner RF & All & 84.4$\pm$0.8  &  90.4$\pm$0.8 \\		
		 Using $\pa'(Y)$ & $\pa'(Y)$ &  \textbf{85.0$\pm$0.8}*		& \textbf{90.6$\pm$0.7} \\
		\hline
	\end{tabular}
	\label{simulation_table}
\end{table}

\begin{table}
	\centering
	\footnotesize
	\caption{A comparison of CCTM-RF and CCTM-SVM with other methods in two groups of data sets. `$+$' indicates that a CCTM method performs better than a corresponding method at 95\% confidence level, `$-$' worse than, and `o' no difference. CCTM methods perform better in 17 cases, and worse in 3 cases. }
	\begin{tabular}[width  = \textwidth]{ | c | c | c | c| c|}
	\hline 
		 \multirow{2}{*}{}  & \multicolumn{2}{c|}{CCTM-RF} & \multicolumn{2}{c|}{CCTM-SVM} \\
\hline
	Method&Group1&Group2&Group1&Group2\\	
	\hline
Causal Tree& $+$ & $+$ & $+$ & $+$    \\
\hline

RespLR-Gen& $+$ & $+$ & $+$ & o   \\

\hline

RespSVM-Linear& $+$ & $-$ & $+$ & o   \\

\hline

t-Stats Tree& $+$ & $+$ & $+$ & $+$  \\	
\hline
Uplift CCIF& $+$ & o & $+$ & o  \\

\hline
Uplift RF& $+$ & o & $+$ & o  \\
\hline

X-Learner RF& o & $-$ & o & $-$ \\

\hline

	\end{tabular}
\label{simulation_comparison}
\end{table}

\subsection{Evaluation on real world data sets}
\label{sec_Exp-Realworld}

We evaluate CCTM-RF and CCTM-SVM by benchmarking with those uplift (causal heterogeneity) modelling methods in two real world data sets, Hillstrom~\cite{Hillstrom_EmailData} and Twins~\cite{Almond2005_TwinData}. In this evaluation, the estimated uplifts (conditional causal effects) are not dichotomised and the capability for identifying subgroups with largest uplifts (causal effects) is evaluated.

Hillstrom contains 42613 customer records from an email marketing campaign collected for an uplift modelling challenge~\cite{Hillstrom_EmailData}. 
Half of these customers were randomly chosen to receive an advertisement email targeting male users, and the other half of the customers served as a control group.  
There are 7 pretreatment variables describing customers. The outcome is whether a customer visits the website. MMPC finds three parent variables for the outcome.  

Twins~\cite{Almond2005_TwinData}
consists of 4821 samples of twin births (with birth weight $<2$ Kg and having no missing values) in the USA between 1989 and 1991. Each record contains 40 pretreatment variables. describing biological parents, pregnancy and information about the birth. 
Treatment $T\! = \!1$ indicates the heavier one in the twins and $T\! =\! 0$ indicates the lighter one. The outcome is the mortality of a child after one year. MMPC finds 4 parent variables for the outcome. 
\begin{figure*}[!t]
	\begin{tabular}{ll}
	\includegraphics[width=0.45\textwidth]{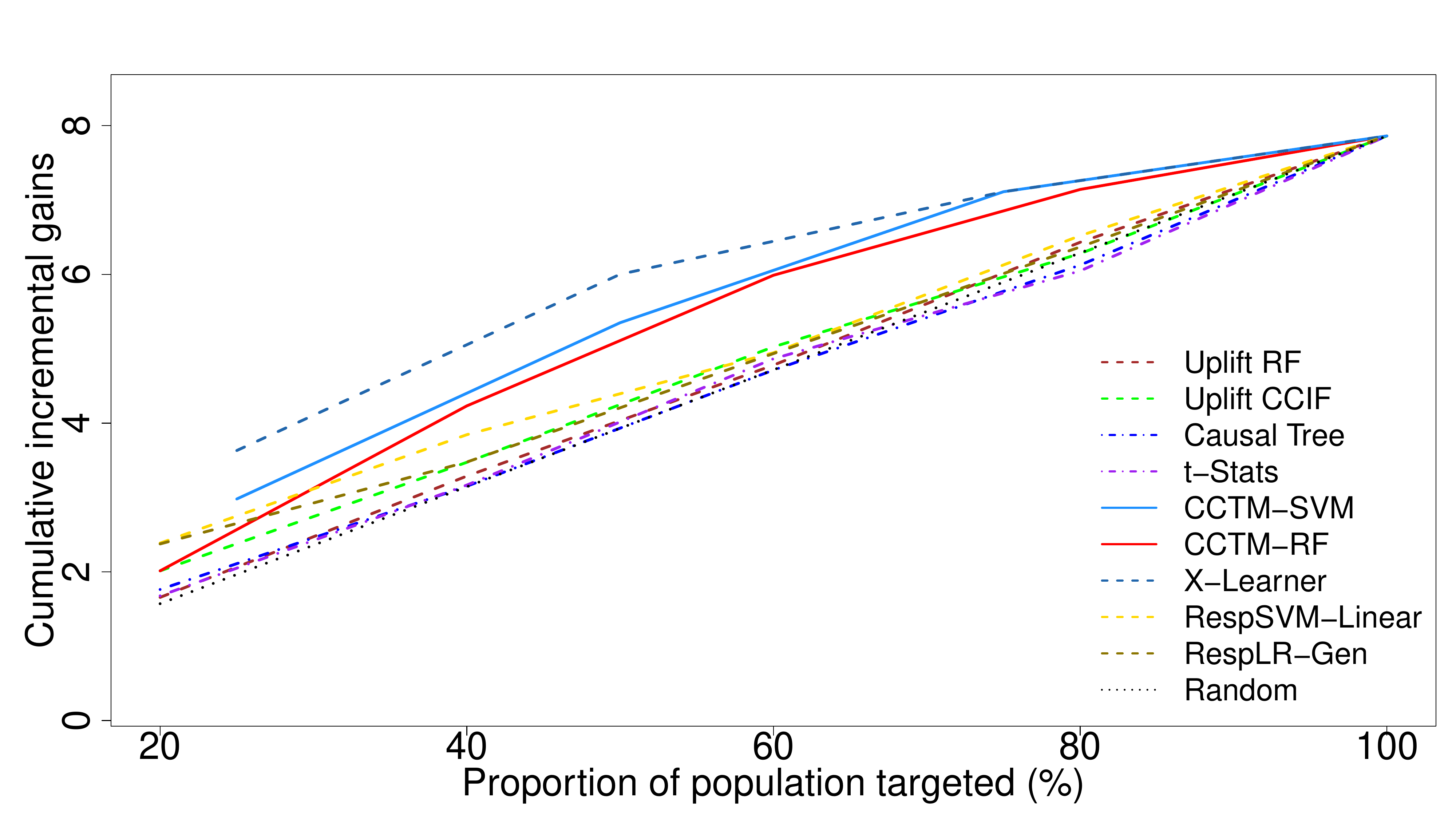} &
	\includegraphics[width=0.45\textwidth]{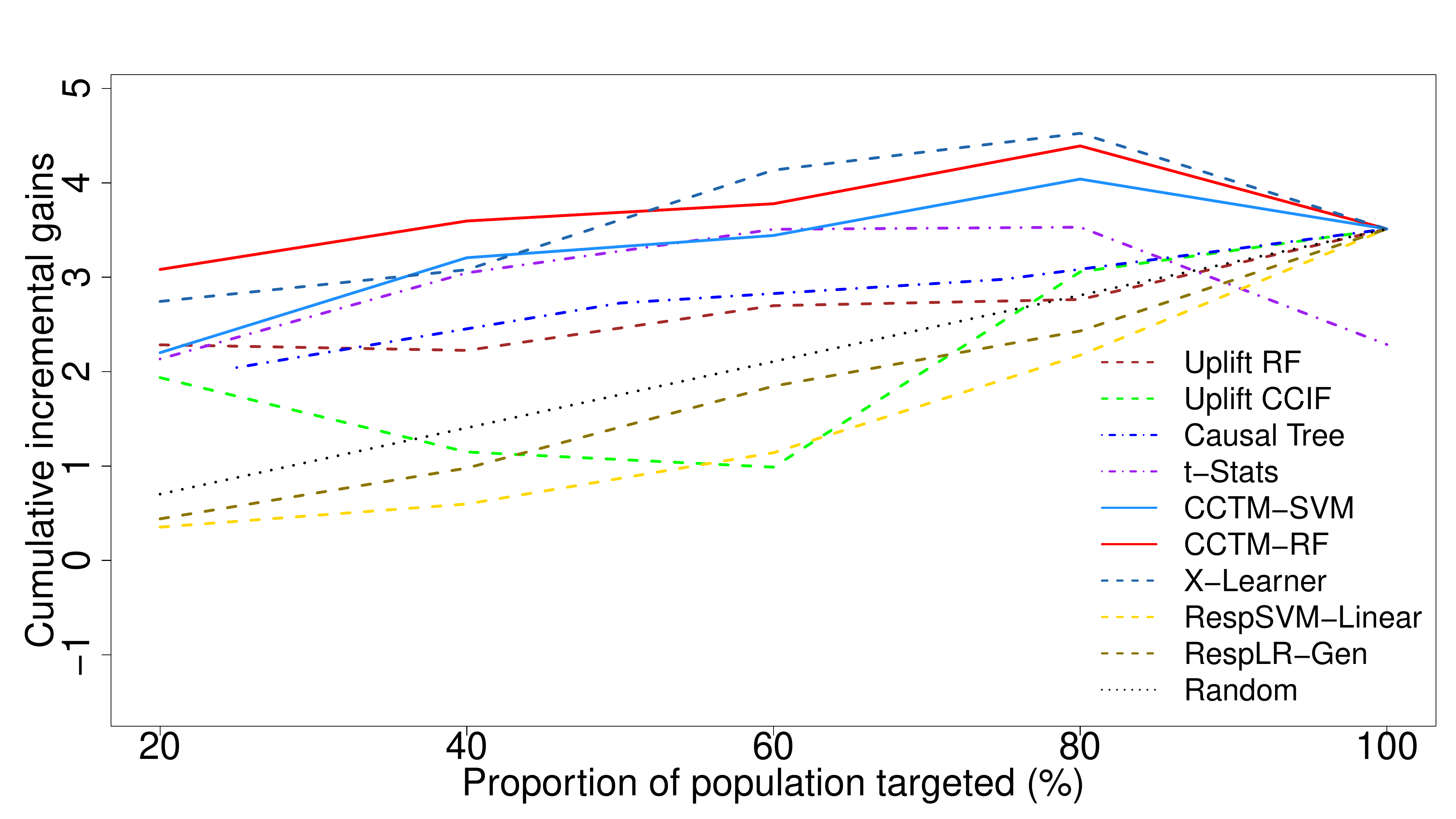} \\
	\end{tabular}
	\caption{The Qini curves of different methods on the two real world data sets (Left: Hillstrom, Right: Twins). Blue and red solid lines: the proposed methods; Dashed lines: other methods; Black dotted line: a random model. }
	\label{fig-QiniCurve}
\end{figure*}

Since there are not ground truth uplifts (or conditional causal effects), we use the Qini curve \cite{Radcliffe2007_Qini}, a widely used metric for uplift modelling to compare the algorithms. Qini coefficient is calculated as $n_{Y=1,T=1} - (n_{Y=1,T=0}\cdot n_{T=1})/(n_{T=0})$, where $n_{Y=1,T=1}$ and $n_{Y=1,T=0}$ are the numbers of positive outcomes in the treatment ($T=1$) and control ($T=0$) groups respectively, and $n_{T=1}$ and $n_{T=0}$ are the total numbers of samples in the treatment and control groups respectively. 
Qini curve shows the cumulative amount of the uplift as a function of the proportion of test samples treated in descending order of predicted uplifts (conditional causal effects). The larger the area a curve covers, the better the corresponding method is. A 10 fold cross validation is used to obtain the Qini curve for each method, and parameter selection is as discussed before.    

Figure~\ref{fig-QiniCurve} shows that the CCTM instantiations, CCTM-SVM and CCTM-RF, achieve competitive performance with other compared methods, ranked the second and third with the Hillstrom data set, and the third and first with the Twins data set. This means that CCTM can also work for uplift (causal heterogeneity) modelling where continuous uplifts (conditional causal effects) are used instead of dichotomised uplifts (conditional causal effects). 

In sum, the experimental results demonstrate that the proposed framework works with off-the-shelf methods. Parent discovery is essential in our framework and Section~\ref{sec_causalEffectestimation} shows that it is helpful for existing uplift (causal heterogeneity) modelling methods too. For both causal classification and uplift (causal heterogeneity) modelling tasks, the two CCTM instantiations perform competitively in comparison with other methods. Note that, the purpose of these experiments are not to demonstrate which methods are better than the others, but to show that the proposed framework can be instantiated using off-the-shelf methods and can achieve competitive performance with other existing methods. The framework is principally correct when assuming causal sufficiency, and provides a means for users to implement any causal classification methods fitting their applications.    

We are aware of the criticisms on two model approach in~\cite{Significance-BasedUpliftTrees2011}, but the recent surveys and evaluations~\cite{Gutierrez2017_CausalUpliftReview,DevriendtSurvey2018,GubelaEvaluation2019} have shown that a two model method performs competitively with other methods. In some cases, a two model method performs the best. However, the conditions for using a two model method have not be discussed in the previous work, and what we have done in this paper has filled in the gap. The proposed framework supports a number of choices of uplift (causal heterogeneity) modelling methods from the off-the-shelf supervised methods for an application, and users can choose the most suitable one for their application.  

Parent identification is a crucial step for the CCTM framework. If other variables are pretreatment variables and a data set is reasonably large, in our experience, parent discovery is quite accurate. Furthermore, Corollary~\ref{corol-directcases} enables domain experts to review the automatically discovered parents. In case of missing parents or inclusion of false parents, they have different impacts on the causal classification models. In the case of missing parents, some confounding variables may be missed, and this will lead to a bias in conditional causal effect estimation. When false parents are included, false inclusion may introduce high variance in estimation. Both lead to inaccurate models in causal classification.

\section{Related work}

Causal classification is closely related to causal effect estimation and causal effect heterogeneity. The potential outcome model~\cite{ImbensRubin2015_Book} and causal graphical models~\cite{Pearl2009_Book} are two major frameworks for causal effect estimation. 
Causal effect heterogeneity is modelled by conditional average causal effects as the causal effects vary in subpopulations. Su et al. \cite{Su2009_RecursivePartitioning} used recursive partitioning to construct the interaction tree for causal effect estimation in subgroups. Foster et al. \cite{Foster2011_Subgroup} introduced the virtual twins method to define subgroups with enhanced causal effects. In~\cite{WagerAthey2018_RF}, random forest was used to predict the probability of an outcome given a set of covariates and CART was used to find a small set of covariates strongly correlated with the treatment to define the subgroups. Dudik et al. \cite{Dudik2011} developed an optimal decision making approach via the technique of Doubly Robust estimation. Athey  et al. \cite{AtheyImbens2016_PNAS} built the Causal Tree to find the subpopulations with heterogeneous causal effects. An X-Learner method~\cite{kunzel2019metalearners} was proposed for causal heterogeneity modelling with unbalanced treated and control samples. All the methods assume a data set with a known covariate set. Recently, several algorithms have also been proposed to estimate conditional average causal effects using neural networks \cite{Shalit2016,Louizos2017_Twin,Kuenzel2018}, and to estimate individual causal effects in networked observational data~\cite{IndividualCENetworkdata2020,CounterfactualEvaluationNetworkeddata2020}.

Covariate selection is essential for causal effect estimation. Covariate set renders the treatment and the outcome to satisfy the ignorability~\cite{ImbensRubin2015_Book} or unfoundedness assumption. Unlike in an experiment where covariates are normally selected by domain experts, data driven covariate selection is very challenging since ignorability is impossible to be tested in data. Data driven methods use the backdoor criterion~\cite{Pearl2009_Book} to identify a covariate set, either based on a causal graph created using domain knowledge or learned from data. VanderWeele and Shpitser~\cite{VanderWeele-NewCriteria} linked the conditional ignorability with the backdoor criterion. de Luna et al.~\cite{de2011covariate} and Entner et al.~\cite{entner2013data} have proposed methods to find covariate sets using conditional independence test. Maathuis and Colombo~\cite{maathuis2015generalized} generalised the backdoor criterion for data without causal sufficiency. 

Uplift modelling is another line of work for estimating conditional causal effects, mainly in marketing research where data collection is through some experimental designs. 
Causal effect has not been mentioned in uplift modelling, but fundamentally, uplift modelling is a type of causal inference~\cite{Gutierrez2017_CausalUpliftReview,Fernandez2018_CausalClassification}. The first proposal of uplift modelling is by Radcliffe and Surry~\cite{Radcliffe1999_DifferentialResponseAnalysis}, Hansotia~\cite{Hansotia2002} and Lo~\cite{Lo2002_TrueLiftModel}. In the well designed experimental data set, Rzepakowski and Jaroszewicz adapted decision trees for uplift modelling~\cite{Rzepakowski2010_DTuplift,Rzepakowski2012_DTforUpliftModel}. Similar adaptions have extended to Bayesian networks~\cite{Nassif2012_BNforUpliftModel} and SVMs~\cite{Nassif2013_SVMforUpliftModel}.  In a similar fashion to the CATE estimation literature, ensemble methods have been introduced to model uplift using a forest of uplift modeling trees \cite{Guelman2015_UpliftRF}. A special case of transformed outcome method has also been introduced to uplift modeling using off-the-shelf estimators directly on the transformed outcomes \cite{Jaskowski2012_Uplift}. Uplift modelling has recently been linked to causal effect heterogeneity modelling~\cite{Gutierrez2017_CausalUpliftReview,Fernandez2018_CausalClassification}, but no unified algorithmic framework has been presented.  Some method surveys and comparisons can be found~\cite{KaneTrulyResponsiveModel2014,DevriendtSurvey2018,GubelaEvaluation2019}. 

Causal classification is a concept used by Fermandez and Provost~\cite{Fernandez2018_CausalClassification}, and authors reported a comparative theoretical analysis between normal classification and causal classification. Authors in~\cite{Fernandez2018_CausalClassification} claim that in some conditions such as, when the outcome is rare or difficult for predict, or the causal effect is small, normal classification performs as good as causal classification. This claim should be assessed in real world data sets.  Kallus~\cite{kallusCausalclassifiction19} presents discriminative and generative algorithms for causal classification based on the monotonicity assumption. In the binary treatment and outcome case, the negative responses are ignored based on the monotonicity assumption. Kallus uses propensity scores to weight the outcomes and convert the causal classification problems to normal discriminative and generative problems. The monotonicity assumption might be strong for some applications.      



\section{Conclusion}

This paper presents a general framework for causal classification, which generalises both uplift and causal heterogeneity models. We have developed a theorem which identifies the conditions for causal classification in observational data and links uplift modelling with causal heterogeneity modelling. The theorem enables a general framework for causal classification using off-the-shelf machine learning methods. We have shown that our theorem improves existing uplift modelling and causal effect heterogeneity modelling methods for better causal effect estimation and  our algorithms have competitive performance compared to other uplift modelling and causal heterogeneity modelling methods in synthetic and real world data sets. 

The causal sufficiency assumption is strong for many real world applications and we will study how to relax the assumption in future. Modelling interactions between variables should be an interesting direction to explore in future.           

\section*{Conflict of interest}
On behalf of all authors, the corresponding author states that there is no conflict of interest. 

\section*{Acknowledgement}
This work has been partially supported by ARC Discovery Projects grant DP170101306.



\end{document}